# COMPUTER VISION FOR A CAMEL-VEHICLE COLLISION MITIGATION SYSTEM

Khalid AlNujaidi and Ghadah AlHabib

Computer Science Department, Prince Mohammad Bin Fahd University,
Khobar, Saudi Arabia.

*ABSTRACT*

*As the population grows and more land is being used for urbanization, ecosystems are disrupted by our roads and cars. This expansion of infrastructure cuts through wildlife territories, leading to many instances of Wildlife-Vehicle Collision (WVC). These instances of WVC are a global issue that is having a global socio-economic impact, resulting in billions of dollars in property damage and, at times, fatalities for vehicle occupants. In Saudi Arabia, this issue is similar, with instances of Camel-Vehicle Collision (CVC) being particularly deadly due to the large size of camels, which results in a 25% fatality rate [4]. The focus of this work is to test different object detection models on the task of detecting camels on the road. The Deep Learning (DL) object detection models used in the experiments are: CenterNet, EfficientDet, Faster R-CNN, and SSD. Results of the experiments show that CenterNet performed the best in terms of accuracy and was the most efficient in training. In the future, the plan is to expand on this work by developing a system to make countryside roads safer.*

## 1. INTRODUCTION

Wildlife-vehicle collision is a global issue. This issue presents itself in a similar manner through the involvement of different species throughout the continents around the world. In North America, as well as some parts of Europe, it is deer that are the main cause of wildlife-related traffic accidents; kangaroos in Australia; and camels in the Middle East and North African regions [1]. This global issue of Wildlife-Vehicle Collision (WVC) has been continuously on the rise throughout the past century, this being a consequence of the human population increase, urbanization of countryside, and the pavement of new roads and highways. This issue is expected to continue to grow as fast as the human population continues to grow.

Occurrences of WVCs result in a wide range of losses such as property damage, disturbance of the ecosystem, and morbidity and mortality for those involved. It is recorded that in the United States, within a year, a total of 247,000 WVCs occurred involving deer, resulting in 200 human fatalities and $1.1bn in property damage [2]. Deer being on the smaller side of wild animals, it is with larger animals that harsher consequences arise. Moose, being a larger animal, constitute more fatal outcomes for the occupants of the vehicles. In Sweden, 4092 WVCs involving moose occurred in a year, resulting in around a 5% fatality rate, accompanied by other serious permanent injuries and a large amount of property damage [3]. A larger and one of the most fatal animals to be involved in WVCs is the camel. Camel-Vehicle Collision (CVC) is considered extremely fatal due to the physical nature of the animal, as in almost all cases the animal tends to

fall through the windshield of the colliding vehicle. CVCs result in as high as a 25% fatality rate [4]. 22,897 camel-related accidents have been recorded over the years 2015-2018 [5]. It is unfortunate, however, that detailed data on CVC regarding the frequency of occurrences, location, and extent of property damage is not readily available. However, a simple web search on the topic always results in relatively recent news headlines with a new occurrence of CVC, along with the graphic details and imagery associated with these accidents.

There has been, and continues to be, effort put into deploying countermeasures to reduce WVC. The most commonly deployed tactics currently in place are conventional ones such as fencing and reflective warning signs. As effective as they may have been, signs can go unnoticed by drivers, and animals have found ways through placed fences [6]. These methods require significant funds and labour to set up and maintain. Therefore, it is necessary to consider developing smarter, technologically advanced, and autonomous methods to act as countermeasures to WVC. Just as it has become very common around the world to use sensors and computer vision technologies to assist in the enforcement of traffic violations, the same can be done to mitigate WVCs and, as the focus of this work, CVCs.

The focus of this work will be to evaluate different state-of-the-art object detection algorithms to serve as a base for a CVC avoidance system. The vision of the work is to further develop an autonomous mechanism that makes countryside roads safer. As has been covered so far in the introduction, the issue of animals colliding with vehicles is a global issue. There are solutions, though they can be very costly and can be improved upon to utilize newer technologies to achieve better results. In the next section, there will be a review of literature related to solutions for both global WVCs and local CVCs. Furthermore, a section will be dedicated to the discussion of our proposed system and the methodologies used. Lastly, there will be a section to review the results, summarize the work, and express the vision for future development.

## 2. RELATED WORK

In [7], the researchers propose an IoT solution for a Camel-Vehicle Collision avoidance system. Their solution consists of two parts: a detection system and an alarm system. The detection system is based on an Omni-directional radar that is responsible for detecting movement and uploading it to the cloud to be analysed. The alarm system reads the data from the cloud and proceeds to turn on/off the alarm signs and horns. In addition, they propose installing a wireless chip that controls nearby vehicles to slow down their speed. This wireless chip is controlled through the alarm system.

In [8], the authors propose a camel crossing alert and tracking system. The method solution they provide is meant to allow camel owners to track their cattle, as well as provide a warning to drivers when the camels approach the road. The system is supposed to be able to detect with a range of 18 KM. This is made possible through the tracking of the geolocation of the camels using LoRaWAN. This is made possible by having collars on the camel's neck that enable a connection to control units spread out on the roads. The alarm system consists of flashing road signs.

In [9], a similar LoRaWAN and GPS system is proposed to combat the issue of camel-vehicle collisions. The method they present is to implant LoRa sensors in the skin of the animals. These sensors are connected to nodes and sub-nodes that create different caution zones along the roads. Once a camel is detected by a node, a signal will be sent to the base station. Each of the nodes are equipped with a GPS system. The alarm system is composed of a mobile phone application that is connected to the base station and sends a message with caution sounds warning drivers about the location where the camels may cross.

The authors in [10] propose a warning system designed for camel-vehicle collision mitigation. The solution they provide consists of night vision cameras installed on several different vehicles. Upon detection with cameras, a system sends a message containing the geolocation through the use of a cellular network. From the messages, a heatmap is derived showing the probability of camel distribution in the area. In addition to the plot of prediction of the movement of camels and where they will be, this is achieved through the use of a hidden Markov model. The alarm system proposed is a message and alarm based on an app installed and linked to both the camera and central base system.

In [11], a review of several methods used for the purpose of mitigating animal-vehicle collisions is conducted. The researcher goes over statistics of the camels roaming in the Gulf region and discusses what methods have been proposed to mitigate camel-vehicle collisions and their efficiencies. It presents the fact that the most practical and effective method currently used is fencing the roads. The authors discuss the harm of closing off animals from crossing the road. Therefore, they propose a method that improves upon fencing by adding regions to the road that work as automated gates for camels to pass. This is done by attaching a radio collar to the camels. The alarm system is based on a signal being sent to a nearby cellular tower that broadcasts an SMS message to nearby drivers, in addition to warning lights flashing near the gates.

A review of implemented animal-vehicle collision mitigation systems is conducted in [12]. It reviews the rate of accidents that occur in different parts of the world, as well as the types of animals involved. The review also discusses implemented and proposed methods for detecting camels on roads in the Middle East and the systems that address this issue. The authors propose a road-based system that consists of wireless IR sensors on the sides of the road, arranged in clusters and connected to a sink node. Each node has a thermal camera and an ultrasonic sensor. When movement is detected by the sensor, an image is taken with the thermal camera, and the sink node analyses the image. If the analysis indicates the presence of an animal, a flashing red alarm light is triggered on both sides of the road.

Some computer vision and artificial intelligence solutions have also been proposed in other regions of the world. In Australia, the authors in [13] propose a region-based convolutional neural network solution for detecting kangaroos in traffic. They work on creating a vehicle-based framework that warns drivers of oncoming kangaroos. The system is composed of a camera or 3D LIDAR. Because of the lack of annotated data on kangaroo activity in traffic, images were generated using photo-realistic simulation and game engine frameworks.

"Where's The Bear?" is an end-to-end framework that was created in [14] for automating image processing and animal detection through an IoT system. The project consists of three parts: the cloud, the edge, and the sensing. The automatic image processing is done by training a model using Google's TensorFlow and OpenCV technologies. The model was trained using generated images of several different animals based on the existing backgrounds found in the field where the system is deployed. IoT sensing devices (cameras) were deployed over 6000 acres at the UCSB Sedgwick Reserve. The distributed sensors are triggered by motion and capture an image, label it, and send it to the database for further analysis of the reserve.

A deep convolutional neural network (DCNN) wildlife monitoring method was proposed in [15]. The focus of the authors in this research is to improve upon existing methods for analysing and tracking 20 different species. The training model consisted of 1100 images of different animals in their natural habitats, with bounding boxes labelling the region where the animal is present. There were also images of plain nature backgrounds with no animal to increase the diversity of the training set. The model was assessed using trap cameras to gather the images, and an accuracy of 91.4% was achieved

## 3. METHODS

This section will cover the dataset used for the experiments, and the annotation process for labelling the data. Deep learning object detection is used for the experiments, a brief description of how the algorithms works will be covered. The workflow is shown in (Figure 1)

### 3.1. Dataset Details

The dataset consists of 250 of camels (Figure 1) in different contexts. Some of the images contain camels in the desert, captivity, and roaming on the highways. Collection of the images was through several different online resources. A handy Google Chrome extension called 'Download All Images' was used to assist in gathering the images, it automatically compressed and downloaded a zip file of all photos present in a webpage. For the use of the object detection algorithms in these experiments the images required annotations. These annotations are BBs of the locations of the object of interest (i.e., camels) within the image to assist in training a model based on the used algorithms. A handy tool named 'Open Labelling' developed for [16] was used for annotating the images.

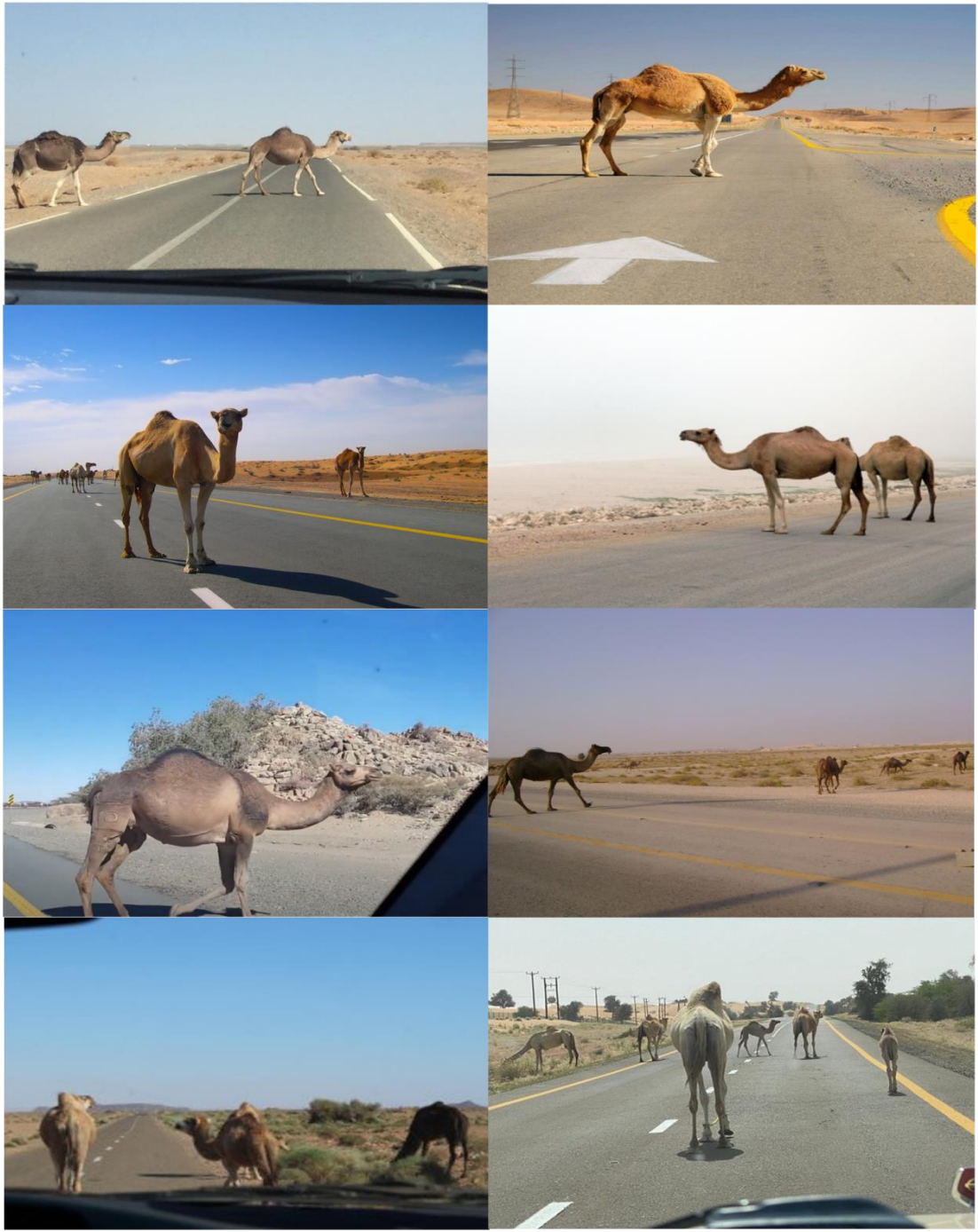

Figure 1. Example images from the dataset

## 3.2 Object Detection

With the advancements of Convolutional Neural Networks (CNN), now computers are not only able to classify what is within an image, computer also is able to localize the object and draw Bounding Box (BB) around it. Only within the last decade, there have been remarkable remarks in regards of developing different object detections algorithms.

The functionality of object detection algorithms is based off those of a CNN. Where there are convolutional layers to extract the features within an image, creating feature maps. Pooling layers to help reduce the dimensionality of the extracted feature maps, this is for the purpose of reducing the computational cost as the CNN becomes deeper. Followed by a fully connected Neural Network for the purpose of classification. With object detection the images are broken down into regions. However, the classification and detection are done in two parts generally after extraction of features through convolutional process (Figure 2). First stage is a probabilistic calculation for classifying what the content of the image within a region. Second stage is application of regression calculations in order to find most optimal BBs. There are a number of different techniques for performing object detection, but most approaches can be grouped into one of two categories: region-based object detection and image classification.

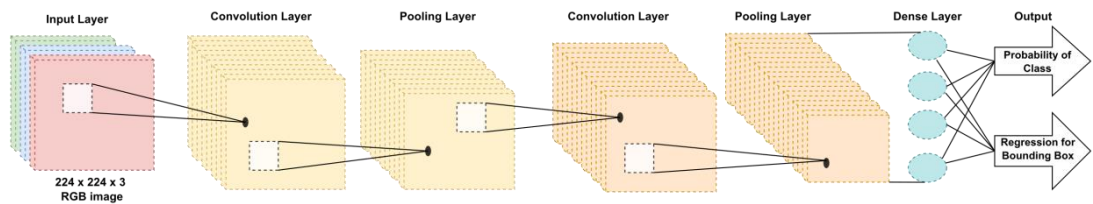

Figure 2. General working of object detection models.

## 4. RESULTS AND DISCUSSION

Four pre-trained models were used for the experiments: CenterNet, EfficientDet, Faster R-CNN, and SSD. First discusses is how the performance of ML models are generally evaluated, and then covering specifically how object detection models are evaluated. After which, the results of the experiments will be shared.

### 4.1. Evaluation Metrices

Object detection models are uniformly evaluated using accuracy of the detection boxes through mean Average Precision (mAP) and mean Average Recall (AR). Intersection over Union (IoU) (Figure 3), Recall, and Precision, are helper metricesused to obtain the desired mAP measurement.

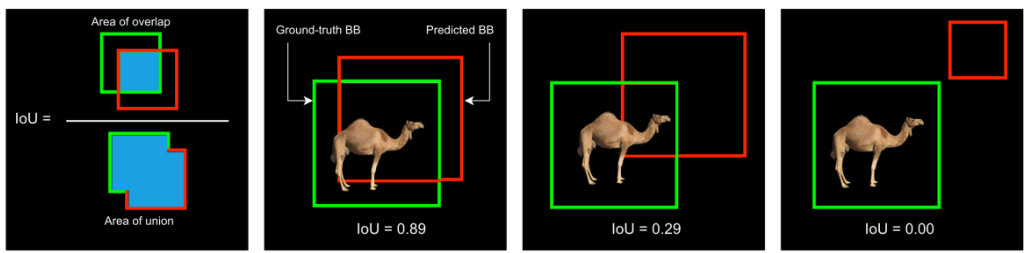

Figure 3. IoU calculation formula. IoU of 0.50:0.95 is considered TP detection.

First, some common evaluation terms and abbreviation widely used in ML model evaluation are the following: True positive (TP), Correct detection made by the model.True Negative (TN), No detection where/when none needed by model. False Positive (FP),incorrect detection made by the model.False Negative (FN), missed detection by model.

International Journal on Cybernetics & Informatics (IJCI) Vol. 12, No.1, February 2023

To reach the desired performance metrics, the procedure required of following the four equations 1-4 are used to find the mAP:

1) $Percision = \frac{TP}{TP+FP}$
2) $Recall = \frac{TP}{TP+FN}$
3) $Average\ Percision = \int_0^1 p_{(r)}\,dr$
4) $mAP = \frac{1}{N}\sum_{i=1}^{N} AP$

Note, *P(r)* being the curve of plotting Precision-Recall, and *N* as number of classes.

### 4.2. Results

All the object detection model used in the experiments are obtained from the TensorFlow 2 Detection Model Zoo repository [17]. The models have been pre-trained on the famous COCO 2017 dataset. The models can be configured to custom datasets through the process of few-shot training. Few-shot learning is a type of machine learning where a model is trained on a small number of examples and is then able to generalize to unseen examples. It is particularly useful in situations where it is difficult or expensive to obtain large amounts of labelled training data, as the model can learn to classify new examples using only a few examples as support.

The models used for the experiments were: CenterNet, EfficientDet, Faster R-CNN, and SSD. All the models were trained on the same computer and using a NVIDIA GeForce GTX 1080 GPU. The accuracy of the models can be seen in the following table (Table 1), as well as visualized in the figure (Figure 2) below.

Table 1. Model performance comparison

| Model | mAP | | | AR | TrainingTime (m) |
|---|---|---|---|---|---|
| | *IoU=0.50* | *IoU=0.75* | *IoU=0.50:0.95* | *IoU=0.50:0.95* | |
| *CenterNet* | 83.4 | 62.7 | 58 | 33.1 | 35.5 |
| *EfficientDet* | 81.7 | 55.4 | 52.6 | 31.8 | 47.4 |
| *Faster R-CNN* | 80.4 | 62.2 | 52.5 | 31.1 | 85.8 |
| *SSD* | 74.8 | 60.2 | 47.9 | 29 | 52.2 |

*All models were trained for 10,000 steps.

Over all, the CenterNet architecture proved to be the best model out of the four. It was able to achieve the highest accuracy, while also being the model that trains the fastest.EfficientDet would in come second place for performance, it has a good general detection rate at IoU=0.50, although it less precise when bound to IoU between 0.50:0.95. As for Faster R-CNN and SSD, either the training time or accuracy hindered their performance, making them less viable then the prior two.

147147

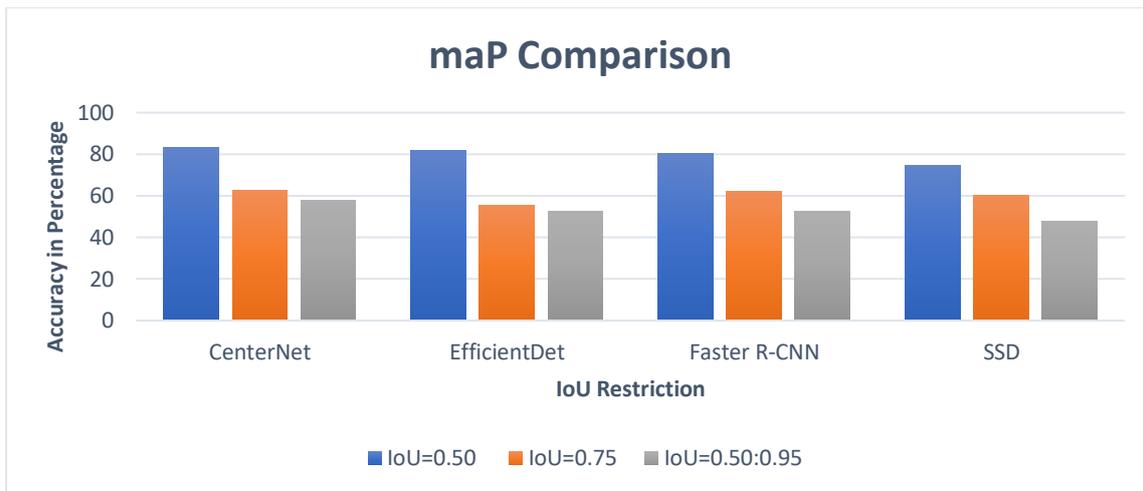

Figure 4. Clustered column graph of models' performances

## 5. CONCLUSION

In conclusion, Wildlife-Vehicle Collisions (WVC) are a global issue that is having a significant socio-economic impact, resulting in billions of dollars in property damage and fatalities. In Saudi Arabia, Camel-Vehicle Collisions (CVC) are particularly deadly due to the large size of camels, which results in a higher fatality rate than other animals.This is a problem that is only expected to grow as the human population grows and more land is used for urbanization. Despite the efforts that have been put to combat this issue, they have not been very effective and are very costly. As computer become smaller and faster and cheaper, it is only making sense to create autonomous systems to combat this issue with warning systems or something of that similar nature. The application of AI and computer vision has proven to be effective in increasing the safety of the roads, with systems such cameras that enforce speed, texting and driving tickets and so on.

As seen in the experiments, with a modest size dataset, satisfactory result has been achieved in detection of camels in different environments. The CenterNet model has proven to be best with a mAP of 58%, and the shortest of training times of only 35.5 minutes. The vision of this work is to then build on these finding and knowledge to them further develop a deployable autonomous system that is able to effectively to both help in reserving natural wildlife ecosystems and prevent property damage.

## 6. THE DATASET

Part of the contribution of this research is to provide a novel type of data that does not exist. A dataset of clean format images and annotated in two styles: Pascal, and YOLO format.
https://www.kaggle.com/datasets/khalidalnujaidi/images-of-camels-annotated-for-object-detection

**AUTHORS**

**Khalid AlNujaidi**. An undergraduate student studying computer science at Prince Mohamed bin Fahd University. Completed a research assistant internship, then continued to be Self-taught in the field of machine learning and artificial intelligence.

**Ghadah AlHabib,** A senior software engineering student at Prince Mohammed bin Fahd University. After completing her bachelor's degree, she will be pursuing a master's degree in artificial intelligence at King Fahd University for Petroleum and Minerals. She has conducted research on Seismic Structures classification using novel features from seismic images. She is currently developing an AI-based innovative education system for the blind. The system is composed of a ring-like device that detects braille and converts it into speech. She is interested in conducting further research in machine learning and computer vision.